\documentclass[review]{elsarticle}

\usepackage{epsfig}
\usepackage{float}
\usepackage{amsmath}
\usepackage{amssymb}
\usepackage{multicol}  
\usepackage{multirow}
\usepackage{subfigure}
\usepackage{booktabs}
\usepackage{lineno,hyperref}

\journal{Journal of \LaTeX\ Templates}

\begin{document}

\begin{frontmatter}

\title{Gram Regularization for Multi-view 3D Shape Retrieval}


%\author{Zhaoqun Li\textsuperscript{1}, Cheng Xu\textsuperscript{3},  Biao Leng\textsuperscript{2,3,4} \fnref{fn2}\\
%\textsuperscript{1}School of Science \& Engineering, The Chinese University of Hong Kong (Shenzhen), Shenzhen, 518172 \\
%\textsuperscript{2}Shenzhen Institute of Beihang University, Shenzhen, 518057\\
%\textsuperscript{3}School of Computer Science \& Engineering, Beihang University, Beijing, 100191 \\
%\textsuperscript{4}Beijing Advanced Innovation Center for Big Data and Brain Computing, Beihang University, Beijing, 100191\\
%\{ lizhaoqun, cxu, lengbiao\}@buaa.edu.cn
%}

\author[add1]{Zhaoqun Li}

\address[add1]{School of Science \& Engineering, The Chinese University of Hong Kong (Shenzhen), Shenzhen, 518172}


\begin{abstract}
How to obtain the desirable representation of a 3D shape is a key challenge in 3D shape retrieval task.
Most existing 3D shape retrieval methods focus on capturing shape representation with different neural network architectures, 
while the learning ability of each layer in the network is neglected. 
A common and tough issue that limits the capacity of the network is overfitting.  
To tackle this, $L_2$ regularization is applied widely in existing deep learning frameworks.
However, the effect on the generalization ability with $L_2$ regularization is limited as it only controls large value in parameters.
To make up the gap, in this paper, we propose a novel regularization term called Gram regularization which reinforces the learning ability of the network by encouraging the weight kernels to extract different information on the corresponding feature map.
By forcing the variance between weight kernels to be large, the regularizer can help to extract discriminative features.
The proposed Gram regularization is data independent and  can converge stably and quickly without bells and whistles.
Moreover, it can be easily plugged into existing off-the-shelf architectures.
%Extensive experimental results on two popular 3D object retrieval benchmarks, 
%ModelNet and ShapeNetCore 55, demonstrate the effectiveness of our method.
Extensive experimental results on the popular 3D object retrieval benchmark ModelNet demonstrate the effectiveness of our method.
\end{abstract}


\end{frontmatter}

\section{Introduction}
3D shape retrieval is a fundamental research problem in 3D shape analysis, which develops rapidly leveraging the ability of Convolution Neural Networks (CNN).
The task is to retrieval 3D shapes in a large-scale dataset and thus requires a discriminative retrieval system to extract high-level shape features.
The retrieval approaches are divided coarsely into two categories, namely model-based methods and view-based methods, depending on the data processed. 
Among the existing 3D shape retrieval methods, view-based methods have achieved the best performance so far.
In view-based 3D shape retrieval, there have been various architectures for processing the rendered images.
In recent years, deep embedding learning over multi-view image sequences of 3D shapes~\cite{johns2016pairwise,su2015multi} has boosted the performance on various 3D shape retrieval benchmark~\cite{wu20153d,chang2015shapenet}.

Recent works focus on capturing the discriminative deep representation of a shape by designing powerful networks.
LFD~\cite{10.1111:1467-8659.00669} is a pioneer view-based method that calculates the similarity between views which is used to obtain the similarity between shapes.
Another popular retrieval algorithm of view-based methods is Multi-view Convolutional Neural Networks (MVCNN)~\cite{su2015multi}
which aggregates multi-views features via a max operation.
This architecture could reduce redundant information between different views and suppress the noisy in some view images.
By exploring the intrinsic hierarchical correlation between views, GVCNN~\cite{feng2018gvcnn} improves the discriminability of features based on MVCNN and achieves better performance.
In a different way,   ~\cite{dai2018siamese} puts forward a bi-direction long short term memory (BiLSTM) model to capture the correlations between different view features and improve the performance for 3D shape recognition and retrieval task,
~\cite{he2018triplet} recently proposes a deep metric learning method called TCL that combines the center loss and triplet loss, 
which achieves state-of-the-art results on various datasets. 

Even though different deep learning architectures for 3D shape retrieval have achieved remarkable performance improvements, 
the learning ability of CNN itself is neglected. 
As CNN prefers to overfit the training set, the results on the testing set are not satisfactory even the training loss converges.
People usually rely on augmenting data to prevent overfitting in the optimization while not pay enough attention to the regularization term of CNN.
The most popular regularization skill in deep learning is $L_2$ term, which is also referred to weight decay in an optimizer.
To some extent, it regularizes the network by reducing the weight norm but its capacity is limited.  
Many research works ~\cite{Chen_2018_Regularizing,Goyal2016Zoneout} propose regularization methods on Recurrent Neural Network (RNN) for various tasks.
These methods can effectively regularize the network by optimizing the distribution of parameters in RNN.
However, they can be employed only in RNN architecture for a specific task.

In order to tackle the aforementioned generalization problem, in this paper,
we propose a novel regularization method called Gram regularization.
Intuitively, our idea lies in that the different weight kernels should extract different information from the feature map and thus obtain more discriminative features.
The inner product operation is considered to form our regularizer that is consistent with the CNN design. 
And our basic strategy is augmenting the variance between different weight kernels leveraging the Gram matrix of a weight group.
The Gram regularization is an extension of $L_2$ regularization with few more calculation cost and it outperforms $L_2$ regularization in retrieval.
Meanwhile, it is data independent and could be applied in different architectures for 3D shape retrieval.
In addition, the proposed regularizer is easy to converge and the training is stable which has no harmful effect on CNN. 
Compared to baseline methods, our Gram regularization brings a large improvement of performance on two representative datasets, ModelNet and ShapeNetCore55.

In summary, our main contributions are as follows.
\begin{itemize}
\item We investigate the regularization methods in 3D shape retrieval and propose a novel regularization term named Gram regularization which can improve the learning ability of CNN in the retrieval task.
\item The Gram regularization is data independent and could be employed in various CNN architectures.
%\item Our method improves different existing methods on both ModelNet and ShapeNetCore55 datasets, which demonstrates that our shape features are more discriminative.
\item Our method improves different existing methods on ModelNet dataset, which demonstrates that our shape features are more discriminative.
\end{itemize}

\section{Related work}
Recently, a large number of research works emerge in 3D shape analysis. 
In particular, benefiting from the generative power of CNN, the methods based on CNN have achieved impressive performance. 
These approaches could be roughly divided into two categories, model-based methods and view-based methods in terms of different raw 3D data that are leveraged.
In this section, we mainly introduce the approaches of 3D shape retrieval using deep learning models and emphasize different schemes of leveraging 2D views' information in view-based methods.

\subsection{Model-based methods}
Model-based methods directly extract the features from the raw 3D representations such as polygon meshes, point cloud and voxel grid, which consist of 3D geometric information. 

Methods leveraging polygon mesh~\cite{xie2017deepshape,boscaini2016learning} is popular for 3D shape retrieval.
~\cite{7559748} proposed a method which is related to unsupervised 3D local feature learning. 
It obtains discriminative shape representation via solving the obstacles in the 3D meshes using a novel circle convolution.
~\cite{7502161} adopts a structure preserving convolution which can simultaneously learn local and global features with structural information. With this strategy, the relation between features is further exploited.
To explore deformable shapes in non-Euclidean domain,~\cite{10.1111:cgf.12693}
proposed localized spectral convolutional network based on localized frequency analysis to obtain robust features.
By encoding the spatial correlation between virtual words, ~\cite{8318683} can generate raw spatial representation with discriminative information that boosts the performance in retrieval.
However, these methods require that the meshes should be smooth.

Voxel-based method~\cite{li2016fpnn,maturana2015voxnet} is a research hotspot as many techniques on analyzing 2D images could be transferred into 3D shape analysis. 
For example,~\cite{wu20153d} proposed 3D ShapeNets that employs CNN to learn feature representations from the 3D voxel grid and achieve impressive performance.
To exploit the abundant information in voxel gird, ~\cite{qi2016volumetric} introduces two volumetric CNN network architectures which consist of part-based classification task and the long anisotropic kernel for long-distance interactions. 
This scheme could take advantage of voxel representation and achieve state-of-the-art performance.
~\cite{8241450} puts forward a novel voxelization strategy which eliminates the effect of rotation and orientation ambiguity on the model surface, further extracting the geometric information.
~\cite{Girdhar16b} proposes TL-embedding network which combines voxels and the corresponding images via two components and could be applied in many tasks.
Although voxel-based methods can exploit the structure information in 3D shapes, they suffer from heavy computational cost which also constrains their ability.

Point cloud is an important role of 3D representation data which can be generated by radar and scanner.
PointNet~\cite{qi2017pointnet,qi2017pointnet++} propose a particular CNN to deal with point cloud.
They use spatial transformation network and symmetric operation to solve the problem of disorder point, obtaining rotation-robust shape features.
To capture local patterns, ~\cite{shen2018mining} propose kernel correlation and graph pooling which focuses on local 3D geometric structures and local high-dimensional feature structures respectively, improving the performance of PointNet.
Although point-based methods can effectively interpret the geometric characteristics of 3D shapes, their performances are limited by the noise in the 3D shapes, such as incompleteness or occlusions.

\subsection{View-based methods}
A common scheme of representing a 3D shape is projecting it to a collection of 2D images, which is called view-based methods.
Recently, the view-based methods lead the best retrieval results on different 3D shape datasets.
In order to aggregate rendered images into a model level feature, three types of model are widely used in 3D shape retrieval: 
single-view based model, multi-view based model and RNN based model.

LFD~\cite{10.1111:1467-8659.00669} is the pioneer view-based method which uses the similarity between views to represent corresponding the similarity between shapes.
Another kind of approaches focuses on post-processing. 
GIFT~\cite{bai2016gift} obtains each view feature by using CNN with GPU acceleration and proposed the inverted file algorithm to reduce computation in the retrieval process. 

MVCNN~\cite{su2015multi} is a classic method adopting CNN to extract the shape feature. 
In this method, the rendered images from different views are fed into a CNN to extract view features.
Then the view features are pooled with the element-wise maximum operation to generate the shape feature.
To obtain rotation invariance, DeepPano~\cite{shi2015deeppano} extracts the shape feature via processing constructed panoramic view with CNNs.
By grouping the views according to their intrinsic hierarchical correlation, 
~\cite{feng2018gvcnn} propose a view-group-shape framework (GVCNN) which largely improves the performance on the 3D shape retrieval.
 ~\cite{huang2018learning} develop a local MVCNN shape descriptors, which generates a local descriptor for any point on the shape and can be directly applicable to a wide range of shape analysis tasks.  
Leveraging the advantages of triplet loss and center loss, TCL~\cite{he2018triplet} is proposed for 3D shape retrieval and achieves state-of-the-art performance. As a deep metric learning method, 
it can optimize the distribution of view features by minimizing the intra-class distance and also maximizing the inter-class distance simultaneously. 

RNN based models are widely applied in natural language understanding~\cite{Wang2017Attention} for its ability of processing sequence data. Different from directly leveraging multiple views, ~\cite{Le2017AM} combines CNN and a two-layer LSTM which encodes the correlation between feature maps for fulfilling 3D segmentation task.
As for the 3D shape recognition and retrieval task, ~\cite{dai2018siamese} proposes a bi-direction LSTM model to capture the correlations between different view features and improve the performance.  ~\cite{ma2018learning} also use LSTM to encode the correlations between views while a three-step training method is adopted for better performance.

% ------------------------
% ------------------------
\section{Proposed method}
In a typical CNN, the learnable parameters (weight kernels) are mainly in convolution layers and fully-connected layers. 
LSTM layer is composed of several fully-connected layers and is widely used in 3D shape analysis for processing view sequence. 
In this article, we only discuss these learnable layers for their important roles in the neural network.
The goal of our method is to obtain a discriminative shape representation via optimizing the weight distribution.
To achieve this, we encourage different kernels to focus on different information in the feature map by reducing the correlation between them.
In this section, we illustrate our method and the its application in different deep learning models of 3D shape retrieval.

% ------------------------
\subsection{Spatial weight group}
\begin{figure}
\centering
\includegraphics[width=\linewidth]{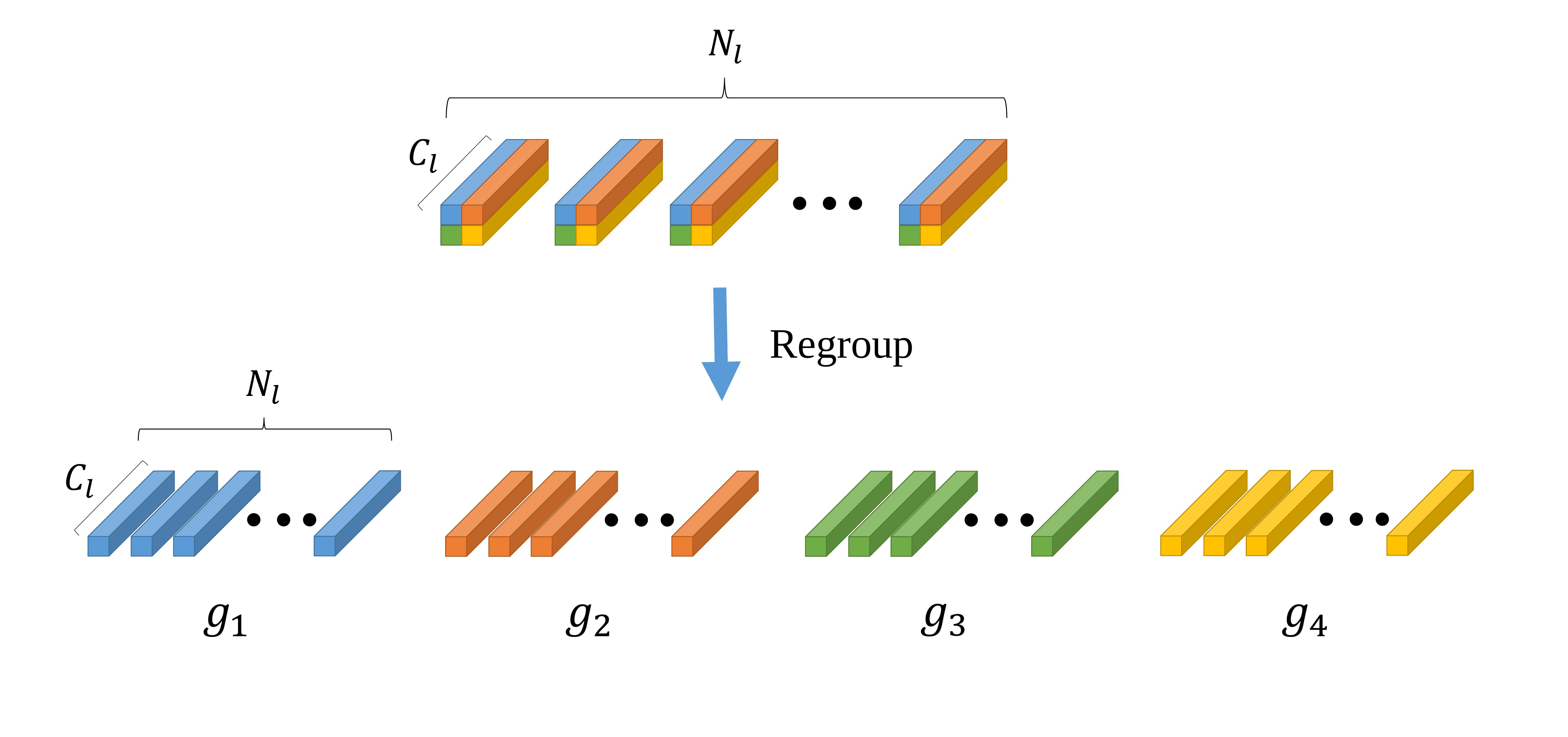}
\caption{The illustration of regrouping process. We take $S_l=4$ as example. 
The weight kernels with $C_l$ channels are divided into different groups depending on their spatial position.}
\label{fig_re}
\end{figure}

For each learnable layer $l$ of the network, 
we suppose that the parameters in $l$ contains $N_l$ weight kernels with dimension  $C_l \times S_l$, 
where $C_l$ is the channels and $S_l$ is the spatial dimension.
To decorrelate the weight kernels, 
our motivation is enforcing different kernels to extract various information in each pixel on the feature map.
So we first regroup the weight by its spatial dimension for the later process.
The regrouping process is shown in Fig. \ref{fig_re}. 
The parameters are divided into $S_l$ spatial weight groups by its spatial position and each group consists of $N_l$ weight vectors with dimension $C_l$.
Formally, let $w(i,j) \in \mathbb{R}^{C_l}$ denote the vector in $i$-th kernel and spatial position $j$. 
The spatial weight group $g_k$ is defined as:
 
\begin{equation}
    g_k = \{ w(i,j) | j=k \}, \quad  k \in \{1, 2, ..., S_l\}
    \label{equation_g}
\end{equation}

From the definition above, we have $g_k \subset \mathbb{R}^{C_l}$ and  $||g_k|| = N_l$.

% ------------------------
\subsection{Kernel Gram matrix}
In deep learning based style transfer methods~\cite{Gatys2015ANA,He2016APG}, a common approach to represent the ``style difference'' between feature vectors is adopting Gram matrix. 
Given a set of feature vector $F=\{f_1, f_2, ..., f_q\}$, 
the Gram matrix $G^{F} \in \mathbb{R}^{q \times q}$ is defined as the inner products between the feature vectors:

\begin{equation}
    G_{i,j}^{F} = f_i^\mathrm{T} \cdot f_j
    \label{equation_G}
\end{equation}
Gram matrix is autocorrelation matrix that represents the style relation between features.
Notice that in the learnable layers of a network, the similarity between a kernel and a feature vector is also encoded by the inner product:

\begin{equation}
    s(w_i,f_j) = w_i^\mathrm{T} \cdot f_j
    \label{equation_ip}
\end{equation}
By this operation, the kernel extracts information from the feature map.
Therefore, the correlation between weight kernels can also be encoded by the inner product.
Here, we employ the Gram matrix in the weight correlation representation:

\begin{equation}
    G_{i,j}^{g_s} = w(i,s)^\mathrm{T} \cdot w(j,s)
    \label{equation_G_s}
\end{equation}
The positively correlated weight kernels contains redundant information, thus we only penalize the postive value in the Gram matrix.
Specifically, we define the Kernel Gram matrix  $K^{l} \in \mathbb{R}^{N_l \times N_l}$ as the sum of filtered Gram matrix of each spatial group: 

\begin{equation}
    K^{l} = \sum^{S_l}_s \max(G^{g_s}, 0)
    \label{equation_kl}
\end{equation}
where $\max(\cdot, \cdot)$ is the element-wise max operation.
The value in the Kernel Gram matrix indicates how much redundant the corresponding weights are. 
In the next section, we will adopt the Kernel Gram matrix to build our regularization term.

% ------------------------
\subsection{Gram regularization}
\begin{figure*}[h]
\centering
\includegraphics[width=\textwidth]{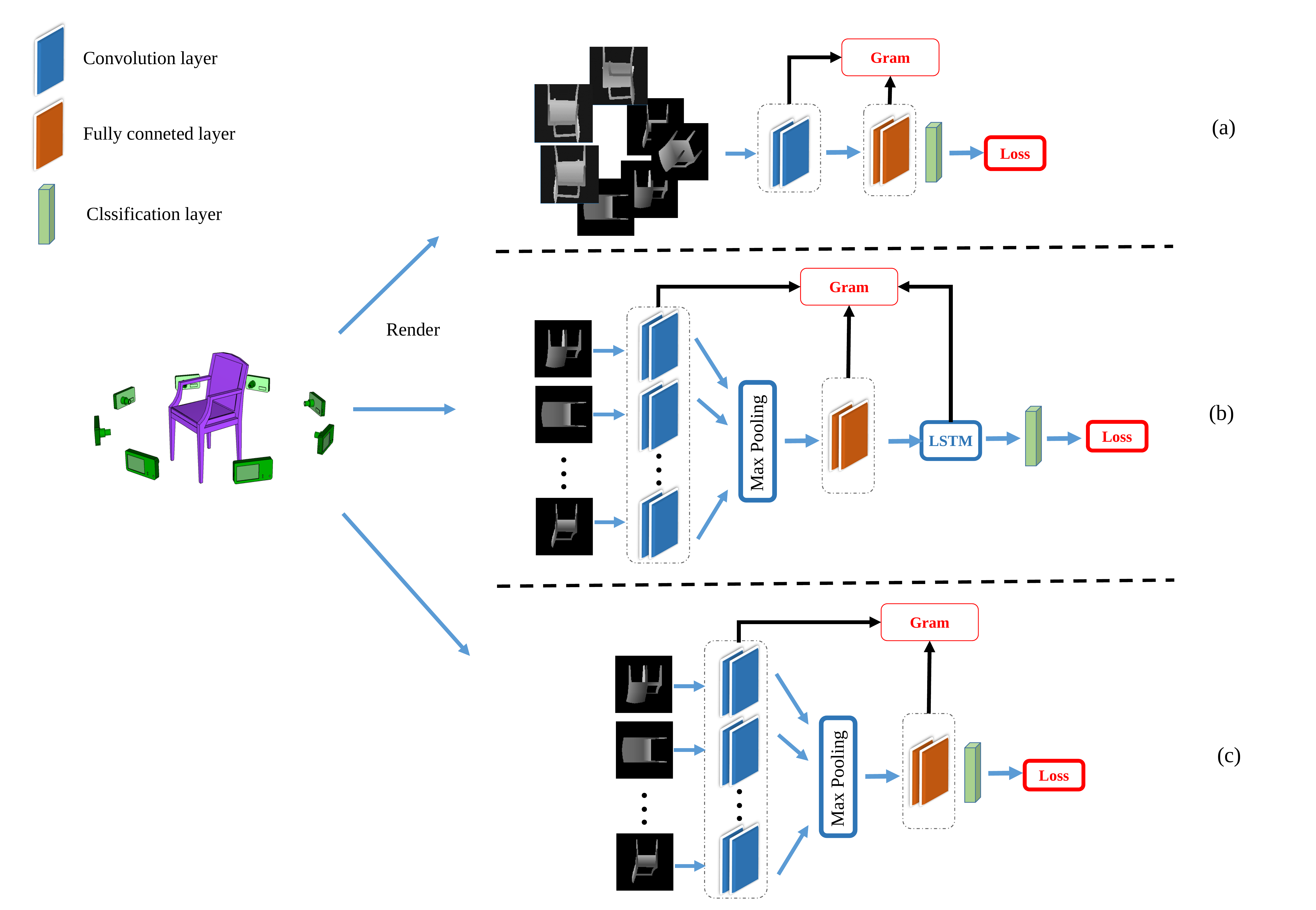}
\caption{The flowcharts of different architectures of deep learning based 3D shape retrieval. 
(a) is View-CNN architecture which learns shape representation by separate rendered images. 
(b) compacts view features of a shape into one shape feature by max operation which is called MVCNN architecture.
(c) is CNN-LSTM architecture that leverages LSTM to process the fixed sequence view features.}
\label{fig_flowchart}
\end{figure*}

$L_2$ regularization is the most popular method to prevent overfitting in deep learning models.
While penalizing the large weight norm, this technique has no direct relation with the weight distribution.
In order to obtain more discriminative shape features in the retrieval task, 
we propose our Gram regularization whcih encourages the weight kernels in one layer to focus on different information.
For a CNN containing $A$ learnable layers, 
let $l_i$ denote the $i$-th learnable layer with parameter set $P_l \in \mathbb{R}^{N_l \times C_l \times S_l}$,
the Gram regulariztion can be expressed:

\begin{equation}
    L^{\prime} = \sum_{l_i}^{L_A} \sum_{(x,y) \in \mathbb{N}^2} K^{l_i}_{x,y}
    \label{equation_gram1}
\end{equation}
It is related to the $L_2$ regularization.
In fact,  $L_2$ can be formulated as:
\begin{equation}
\begin{aligned}
    L_2	= \sum_{l_i}^{L_A} \sum_{w \in P_l}  w^2
    		= \sum_{l_i}^{L_A} \sum_{(x,y), x=y} K^{l_i}_{x,y}
    \label{equation_l2}
\end{aligned}
\end{equation}
From above we can see that the Gram regularization is an extension of $L_2$ regularization.
In order to compare $L_2$ regularization with our method, we separate two terms in Eq. \ref{equation_gram1} by adding trade-off hyper-parameters:

\begin{equation}
\begin{aligned}
    L_{Gram} = &\sum_{l_i}^{L_A} \left[\lambda_1 \sum_{(x,y), x\ne y} K^{l_i}_{x,y}+ \lambda_2 \sum_{(x,y), x=y} K^{l_i}_{x,y} \right] \\
 			    = &\lambda_1 \sum_{l_i}^{L_A} \sum_{(x,y), x\ne y} K^{l_i}_{x,y} + \lambda_2 L_2 
    \label{equation_gram}
\end{aligned}
\end{equation}

In this paper, we adopt Eq. \ref{equation_gram} as our regularization method.
Compared with single $L_2$ term, $L_{Gram}$ also takes the discrepancy between different kernel weights into consideration.
The Gram regularization is also data independent which can be incorporated into existing deep learning structures. 

For the 3D shape retrieval task, softmax loss is often used to guide feature learning over different architectures. 
And thus the total loss can be expressed as:

\begin{equation}
\begin{aligned}
    L_{total} & =  L_{softmax} + L_{Gram} \\
    			& = L_{softmax} + \lambda_1 \sum_{l_i}^{L_A} \sum_{(x,y), x\ne y} K^{l_i}_{x,y} + \lambda_2 L_2 
    \label{equation_lt}
\end{aligned}
\end{equation} 

We will discuss the influence of hyper-parameters on performance in Sec.\ref{sec_dic}.

% ------------------------
\begin{figure}
\centering
\includegraphics[width=\linewidth]{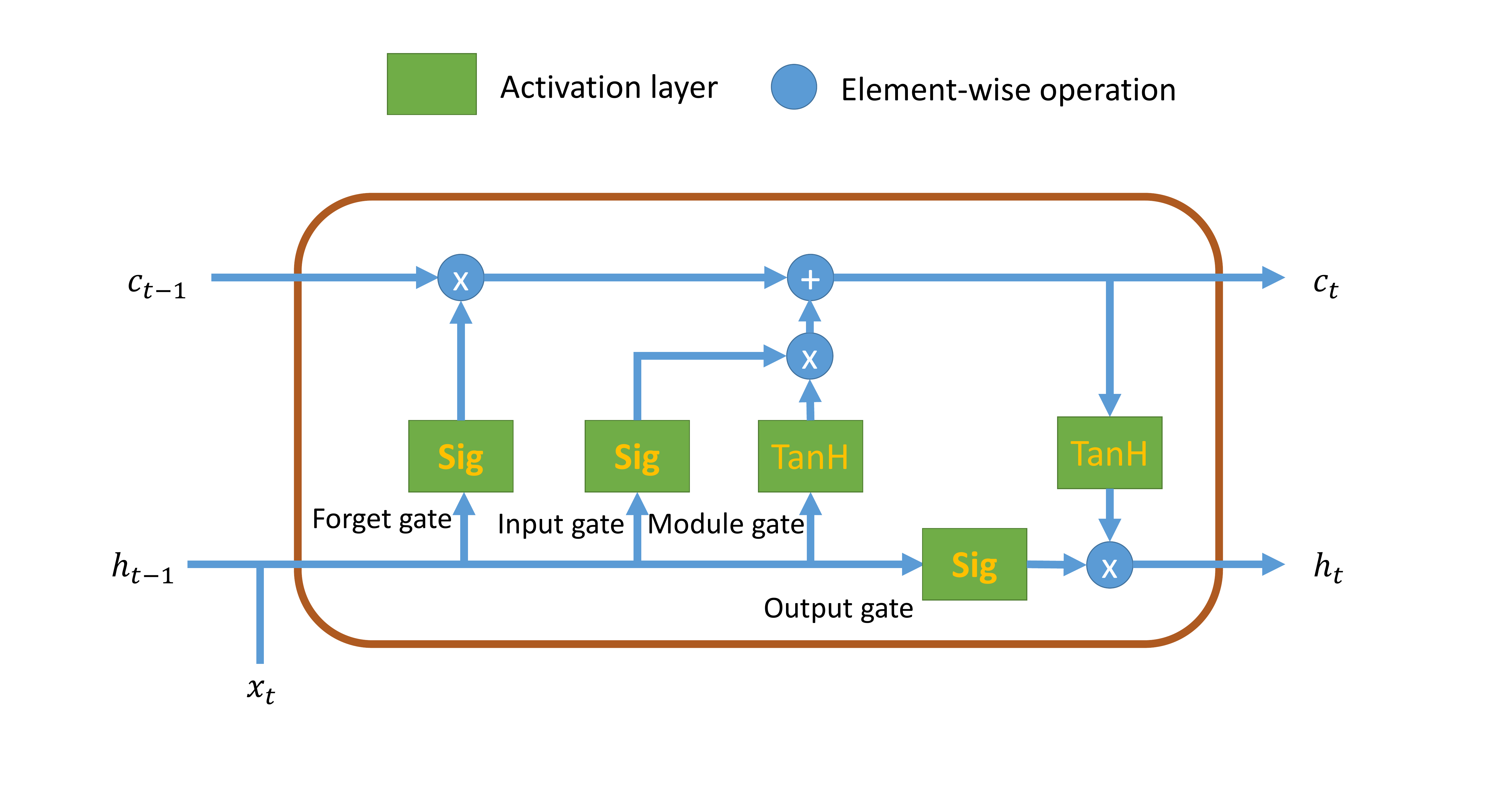}
\caption{The structure of the LSTM cell.  }
\label{fig_lstm}
\end{figure}

\subsection{Application in different architectures}
To investigate the efficiency of Gram regularization, 
we implement our method in three popular deep learning architectures for view-based 3D shape retrieval: View-CNN, MVCNN and CNN-LSTM.
The flowcharts of these architectures are shown in Fig.\ref{fig_flowchart}.
In this paper, we adopt VGG\_M, which includes 5 convolution layers and 3 fully connected layers, as our base network,

\noindent \textbf{View-CNN architecture.}
In view-based methods, the rendered images in different views are put into a CNN to extract the view features.
To extract discriminative features from shapes, the most direct way is training by images which is the strategy of View-CNN architecture.
In the inference, the rendered images of a shape are input into multi-CNNs which share parameters.
Then the shape feature is obtained by simply averaging the view features.

\noindent \textbf{MVCNN architecture.}
Many researchers propose that the intrinsic relation between view features should be explored.
For eliminating redundant information and generating more discriminative shape feature, 
MVCNN~\cite{su2015multi} is proposed that adopts max operation in the middle layer of the network.
Specifically, the element-wise max operation is placed after one layer to aggregate all the middle view features.
Then one feature vector is pooled for the later process. 
We reconduct MVCNN on VGG\_M in the Sec. \ref{sec_exp} and obtain a similar result in our experiment.
There are also many methods in 3D shape retrieval which employ this architecture.  

In our implementation, we place the max operation after \textit{conv5} layer in the MVCNN architecture.
The Gram regularization is adopted in all 7 layers except the last layer as it serves as the classification layer.

\noindent \textbf{CNN-LSTM architecture.}
LSTM plays an important role in solving the problem with sequence data. 
As shown in Fig. \ref{fig_lstm}, each LSTM cell contains four gates that control the information passed into the next cell.
Each gate is composed of two fully connected layers and an activation layer.
Formally, the four gates named input gate $i(t)$, forget gate $f(t)$, output gate $o(t)$ and module gate $g(t)$ are expressed below.

\begin{equation}
\begin{aligned}
g(t) &= \phi(W_{gx}\cdot x(t) + W_{gh}\cdot h(t-1) + b_{g}) \\
i(t) &= \sigma(W_{ix}\cdot x(t) + W_{ih}\cdot h(t-1) + b_{i})   \\
f(t) &= \sigma(W_{fx}\cdot x(t) + W_{fh}\cdot h(t-1) + b_{f})  \\
o(t) &= \sigma(W_{ox}\cdot x(t) + W_{oh}\cdot h(t-1) + b_{o})   \\
s(t) &= g(t) * i(t) + s(t-1) * f(t) \\
h(t) &= \phi(s(t)) * o(t)
\end{aligned}
\end{equation}

Where $W_{gx}, W_{ix}, W_{fx}, W_{ox}, W_{gh}, W_{ih}, W_{fh}, W_{oh}$ are the parameters in a LSTM unit.
Here, $\sigma$ is sigmoid activiation, $\phi$ is tanH activiation and  $*$ denote element-wise multiplication.

In this article, we connect the penultimate layer \textit{fc7} of VGG\_M to the LSTM layer.
In the training and testing process, 
the rendered images of a shape are fed into the network together so that the LSTM layer can discover the correlation between the view features.
Multi-step training strategy~\cite{ma2018learning} is adopted as described in Sec.~\label{sec_exp}.
The Gram regularization is used in each LSTM cell for regularizing its 8 parameters in fully connected layers.

%% ---------------------------------
\subsection{Backpropagation of Gram regularization}
The Gram regularization is data independent and is calculated by weight group, 
so the gradient of each weight is only related to its corresponding weight group.
Here, we first give the formulation of the gradient for a Kernel Gram matrix $K_l$ in layer $l$.
Suppose the layer $l$ contains $N_l$ weight kernels and the spatial position is $S_l$.
$g_k = \{w_i | i=1,2,...,S_l\}$ is the $k$-th weight group (defined in Eq. \ref{equation_g}) in this layer, the gradient of its element is expressed as:

\begin{equation}
\begin{aligned}
    \frac{\partial \sum_{(x,y), x\ne y} K^{l}_{x,y}}{\partial w_i} = \sum_{j=1,j \neq i}^{N_l} \delta(w_i^\mathsf{T} \cdot w_j > 0) \cdot w_j 
    \label{equation_bpkl} 
\end{aligned}
\end{equation}
Where $\delta(cdt)=1$ if $cdt$ is true and $\delta(cdt)=0$ otherwise. 
Eq. \ref{equation_bpkl} shows that the backpropagation is simple which guarantees stability in optimization.
By Eq. \ref{equation_bpkl}, we can obtain the backpropation formulation of $L_{Gram}$:

\begin{equation}
\begin{aligned}
    \frac{\partial L_{Gram}}{\partial w_i} &= \lambda_1 \frac{\partial \sum_{(x,y), x\ne y} K^{l}_{x,y}}{\partial w_i} + \lambda_2 \frac{\partial L_2}{\partial w_i} \\
    &= \lambda_1 \sum_{j=1,j \neq i}^{N_l} \delta(w_i^\mathsf{T} \cdot w_j > 0) \cdot w_j + 2\lambda_2 w_i
    \label{equation_bpg} 
\end{aligned}
\end{equation}   
This formulation is consistent for all the parameters in learnable layers.

%% ------------------------------------
%% ------------------------------------
\section{Experiment}
\label{sec_exp}
%In this section, we evaluate the performance of Gram regularization on two popular 3D shape datasets, ModelNet and ShapeNetCore55.
In this section, we evaluate the performance of Gram regularization on ModelNet dataset.
The experiments are conducted with different CNN structures in view-based 3D shape retrieval and we compare the results with the baseline methods. 
We also investigate the effect of our method on the training process by analyzing the loss variance.
Finally, we discuss the influence of hyper-parameters, $i.e.$ $\lambda_1$ and $\lambda_2$, on the retrieval performance and visualize the retrieval result.

\noindent\textbf{Implementation details.} 
Our experiments are conducted on a server with two Nvidia GTX1080Ti GPUs, an Intel Xeon CPU and 128G RAM.
The algorithm codes are implemented by Pytorch.
For the structure of CNN, we use VGG-M~\cite{chatfield2014return} which has 5 convolution layers (\emph{conv1-5}) and 3 fully-connected layers (\emph{fc6-8}) as the base network in all our experiments. 
The network is pre-trained on ImageNet~\cite{deng2009imagenet}.

Before the training, we render views by Phong reflection in different positions to generate depth images. 
The number of views used in the experiments is 8 by default. 
The size of each image is 224x224 pixels in our experiment.
In the View-CNN architecture, the batch size is 64 and the base learning rate is 1e-2.
For the MVCNN, we place view pooling layer after \emph{conv5} with batch size 20 and base learning rate 1e-3.
As to the CNN-LSTM structure, LSTM layer is added after \emph{fc7} and its output is input into the final classification layer \emph{fc8}.
We train the CNN-LSTM network by multi-step strategy ~\cite{ma2018learning}  with batch size 20. 
First, we initialize the CNN part with the parameters well-trained in the View-CNN since the two architecture have same base network (VGG\_M) before LSTM.
Second, we train the LSTM layer solely by fixing the parameters in the CNN part.
Finally,  the parameters in CNN and LSTM are updated jointly.
The base learning rates for the three steps are 1e-2, 1e-2 and 1e-3 respectively.
In all the experiments, the Gram regulariztion is applied in all layers except for the last classification layer (\emph{fc8}).

We use the stochastic gradient descent (SGD) algorithm with momentum 2e-4 to optimize the total loss.
The learning rate is divided by 10 at the epoch 40.
The total training epochs are 60.
The features extracted for testing are the outputs of the penultimate layer, \emph{i.e. fc7} or LSTM.
The cosine distance is adopted as the evaluation metric.

\begin{table*}
\begin{center}
\scalebox{1}[1]{
\begin{tabular}{|l|c|c|c|cc|}
\hline
Architectures		&Regularization 		& Pretrained	& Fine-tuned			& AUC 			& MAP\\
\hline
View-CNN 		&	$L_2$				& ImageNet1k	& ModeNet40			& 81.32\% 		& 79.93\% 		\\	
CNN-LSTM		&	$L_2$				& ImageNet1k	& ModeNet40			& 82.83\% 		& 81.53\% 		\\
MVCNN 			&	$L_2$				& ImageNet1k	& ModeNet40			& 81.66\% 		& 80.29\% 		\\
MVCNN 			&	$L_2$				& -				& ModeNet40			& 72.76\%  		& 71.05\%		\\

View-CNN 		&	$L_{GRAM}$		& ImageNet1k	& ModeNet40			& 84.44\% 		& 83.19\% 		\\	
CNN-LSTM		&	$L_{GRAM}$		& ImageNet1k	& ModeNet40			& 84.71\% 		& 83.43\% 		\\
MVCNN 			&	$L_{GRAM}$		& ImageNet1k	& ModeNet40			& \textbf{85.02}\% 		& \textbf{83.85}\%		\\
MVCNN 			&	$L_{GRAM}$		& -				& ModeNet40			& 74.65\%  		& 73.02\% 		\\
\hline
\hline
MVCNN 			&	$L_2$				& ImageNet1k 	& ModeNet10			& 86.70\% 		& 85.84\%	\\
MVCNN 			&	$L_2$				& -				& ModeNet10			& 81.75\%  		& 80.54\%	\\

MVCNN 			&	$L_{GRAM}$		& ImageNet1k	& ModeNet10			& \textbf{87.32}\%			& \textbf{88.15}\% 		\\
MVCNN 			&	$L_{GRAM}$		& -				& ModeNet10			& 84.31\%   		& 83.35\%  		\\
\hline
\end{tabular}
}
\end{center}
\caption{The performance comparison of regularization methods on ModelNet.}
\label{tab_m40}
\end{table*}

%\begin{table*}
%\centering
%\begin{tabular}{lccccccccc}
%\hline
%\multirow{2}{*}{Methods} 	 & \multicolumn{3}{c}{Micro} & \multicolumn{3}{c}{Macro}& \multicolumn{3}{c}{Micro + Macro}\\
%\cmidrule(lr){2-4}  \cmidrule(lr){5-7} \cmidrule(lr){8-10} & F-measure & MAP & NDCG & F-measure & MAP & NDCG & F-measure & MAP & NDCG\\
%\hline
%MVCNN 	& 0.612	& 0.734 & 0.843 		&  \textbf{0.416} & 0.662 	& 0.793 	&  \textbf{0.514} & 0.698 & 0.818\\
%Our		&  \textbf{0.619}	&  \textbf{0.807} &  \textbf{0.878} 		& 0.357 &  \textbf{0.667} & \textbf{0.808} 	& 0.488 & 0.737 & \textbf{0.843}\\
%\hline
%\end{tabular}
%\caption{The performance comparison on SHREC16.}
%\label{tab_shrec}
%\end{table*}

%\begin{table}
%\centering
%\scalebox{0.9}[1]{
%\begin{tabular}{lcccccc}
%\hline
%\multirow{2}{*}{Methods} 	 & \multicolumn{3}{c}{Micro} & \multicolumn{3}{c}{Macro} \\
%\cmidrule(lr){2-4}  \cmidrule(lr){5-7} & F1 & MAP & NDCG & F1 & MAP & NDCG \\
%\hline
%MVCNN 	& 61.2	& 73.4 & 84.3 		&  \textbf{41.6} & 66.2 	& 79.3 	\\
%MVCNN+Gram		&  \textbf{61.9}	&  \textbf{80.7} &  \textbf{87.8} 		& 35.7 &  \textbf{66.7} & \textbf{80.8} 	\\
%\hline
%\end{tabular}
%}
%\caption{The retrieval results on SHREC16. Our method also adopts MVCNN and use Gram regularization. }
%\label{tab_shrec}
%\end{table}

\subsection{Retrieval on large-scale 3D datasets}
\noindent\textbf{Dataset.} 
To evaluate the performance of our method, we conduct 3D shape retrieval experiments on  ModelNet dataset~\cite{wu20153d}. This benchmark is a large-sacle 3D CAD model dataset which includes 127,915 3D CAD models classified into 662 categories. 
It has two subsets called ModelNet40 and ModelNet10.
ModelNet40 dataset contains 12,311 models cleaned mannually from 40 categories and ModelNet10 contains
4,899 models from 10 categories.
The number of models in each category is different. 
For our implementation, we follow the same method to split training and test set as described in~\cite{wu20153d}, $i.e.$ randomly select 100 unique models per category from the subset, where 80 models are used for training and the rest for testing. 

The evaluation metrics adpoted in this paper include mean average precision (MAP) and area under curve (AUC)
Refer to~\cite{wu20153d} for their detailed definitions.

\noindent\textbf{Comparison with $L_2$ regularization.} 
In retrieval experiments on ModelNet dataset, 
we choose the generally used $L_2$ regularization as our baseline method.
To demonstrate the efficiency of our method, 
we compare the retrieval performance of our Gram regularizer with classic $L_2$ regularizer in different deep learning structures. 
The experiment results are presented in Tab. \ref{tab_m40}.
For all three structures, our method outperforms $L_2$ regularization by $3.26\%$, $1.90\%$ and $3.56\%$ respectively on ModelNet40 dataset. 
The results show that MVCNN architecture with Gram regularization achieves the best performance.
Under the consideration of performance and computation efficiency, 
we set MVCNN with Gram regularization as our default method in the later experiments.
Then we conduct extra experiments on ModelNet10 using MVCNN architecture and our method also outperforms $L_2$ regularization.
From the table we can observe that, as an extension of $L_2$ regularization, 
our approach imposes extra constraints on the weight kernels which augment the variance between them.

\begin{table}
\begin{center}
\scalebox{1}[1]{
\begin{tabular}{|lc|cc|}
\hline
Methods		                                     		&& AUC			& MAP				\\						
\hline
ShapeNets 									&& 49.94\% 		& 49.23\%     		\\	
DeepPano 									&& 77.63\% 		& 76.81\% 		\\
GIFT 										&& 83.10\% 		& 81.94\% 		\\
MVCNN-su 								&& - 				& 80.20\% 		\\
Siamese CNN-BiLSTM					&& -				& 83.30\%			\\
PANORAMA-NN							&& 87.39\%		& 83.45\%			\\
GVCNN 									&& - 				& 85.70\% 		\\
ATCL 										&& 87.23\% 		& 86.11\% 		\\
CIPLoss									&& 88.21\% 		& 87.22\% 		\\
\hline
MVCNN+Gram 							&& 85.02\% 		& 83.85\%		\\
ATCL+Gram								&& \textbf{88.10\%} 	& \textbf{87.02\%} \\
\hline
\end{tabular}
}
\end{center}
\caption{The performance comparison with state-of-the-arts on ModelNet40.}
\label{tab_sr}
\end{table}

%As to the evaluation in ShapeNetCore55 dataset with a great number of 3D shapes, 
%we also compare the retrieval performance using MVCNN architeture. 
%In this dataset, two labels are provided for the comprehensive comparison and we evaluate the results with regard to Macro and Micro.
%Macro is mainly for providing an unweighted average on the entire database and Micro aims to dispose the influence of different model categories sizes. 
%As shown in Tab. \ref{tab_shrec}, our method brings an improvement compared to the baseline method. 
%Note that the results in the table are based on our implementation with 8 views.

\noindent\textbf{Comparison with state-of-the art methods.} 
As a general regularization method, our regularizer can embed into other retrieval methods and further improve the performance.
To show our method's ability, 
we combine the Gram regularization with ATCL~\cite{li2019Angular} algorithm which achieves state-of-the-art results in 3D shape retrieval.
ATCL is a deep metric learning method adopting MVCNN architecture and uses VGG\_M as backbone in the paper.
To evaluate our method, we just need to replace $L_2$ regularizer by Gram regularizer in ATCL.
We follow~\cite{li2019Angular} for the other hyperparameter settings. 
The comparison results are shown in Tab. \ref{tab_sr}.
We choose 3D ShapeNets~\cite{wu20153d}, DeepPano~\cite{shi2015deeppano}, MVCNN-su~\cite{su2015multi},
PANORAMA-NN~\cite{Sfikas2017Exploiting}, CIPLoss~\cite{li2019Rethinking}, 
Siamese CNN-BiLSTM~\cite{dai2018siamese}, GVCNN~\cite{feng2018gvcnn} and GIFT~\cite{bai2016gift} methods for comparison. 
Compared to ATCL, our regularization method boosts the MAP by $0.91\%$ and achieves the best performance on ModelNet40.

\begin{figure}
\centering
\includegraphics[width=\linewidth]{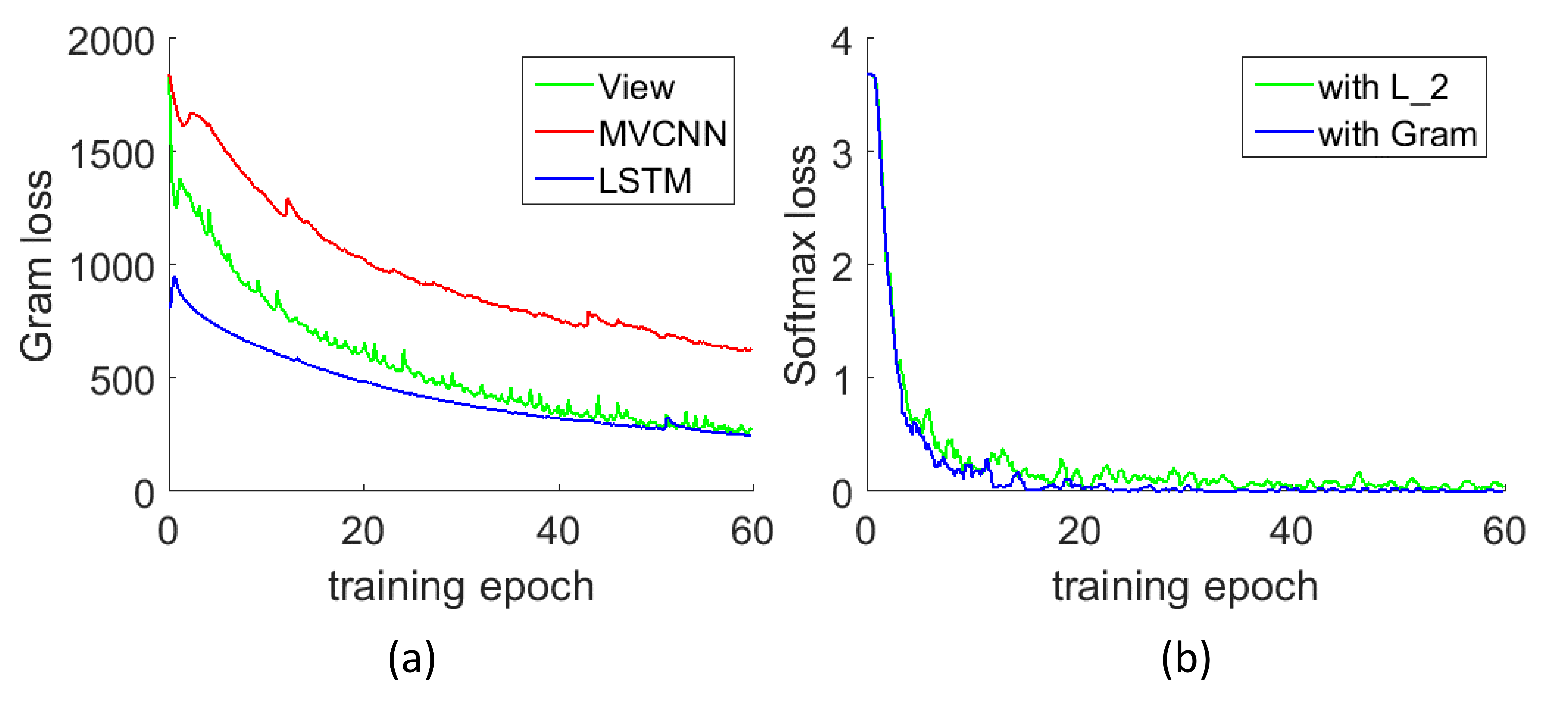}
\caption{Learning curves. (a) plots the drop of regularization loss and (b) depicts the variance of softmax loss with tow regularization terms.}
\label{fig_lc}
\end{figure}

\noindent \textbf{Learning curves.}
The regularization term works as a loss function in CNN and there two major two concerns in the optimization.
First, the regularization itself should converge stably.
Then, the Gram regularization should not influence the drop of target loss, $i.e.$ softmax loss, in the optimization procedure.
To demonstrate the stability of Gram regularizer, we conduct experiments under softmax loss supervision and plot the regularization loss value in Fig. \ref{fig_lc}(a).
The three curves represent the loss variance with three architectures: View-CNN, MVCNN and CNN-LSTM.
From the figure we can see that the Gram regularization drop stably which guarantees stability in the optimization.
Fig. \ref{fig_lc}(b) depicts the influence of Gram regularization on the target loss function.
We set $L2$ regularizer as the baseline method in this comparison experiment.
Compared to the $L_2$ method, our regularization won't harm the drop of target loss but in fact help the softmax loss converge more quickly.

%%%%%%%%%%%%%%%%%%%%%%%%%%%%%%%%%%%%%%%%%%%%%%%%%%%%%%%%%%%%%%%%%%%%%%
%%%%%%%%%%%%%%%%%%%%%%%%%%%%%%%%%%%%%%%%%%%%%%%%%%%%%%%%%%%%%%%%%%%%%%
\subsection{Discussion}
\label{sec_dic}

\begin{table}
\begin{center}
\scalebox{1}[1]{
\begin{tabular}{|l|c|cc|}
\hline
$\lambda_1$		&$\lambda_2$		& AUC 			& MAP\\
\hline
	0				&	1e-4			& 81.66\% 		& 80.29\% 		\\
	0				&	1e-3			& 81.12\% 		& 79.93\% 		\\
\hline
	1e-4			&	1e-4			& 83.75\% 		& 82.67\% 		\\	
	1e-3			&	1e-3			& 84.47\% 		& 83.30\% 		\\
	1e-3			&	1e-4			& \textbf{85.02}\% 		& \textbf{83.85}\%		\\
\hline
\end{tabular}
}
\end{center}
\caption{The performance comparison on ModelNet40 with different hyper-parameter settings.}
\label{tab_hp}
\end{table}

\noindent \textbf{Hyper-parameter analysis.}
There two hyper-parameters in our method, 
namely $\lambda_1$ and $\lambda_2$ which are loss weights of $L_2$ regularization and Gram regularization in the optimization. 
We conduct experiments with different values of hyper-parameters to observe the influence on the final performance.
The experiment results are shown in Tab. \ref{tab_hp}.
The first two lines where $\lambda_1=0$ is our baseline methods.
From the table we can see that much more regularization will be harmful to the network, 
the ideal value for $\lambda_1$ and $\lambda_2$ is 1e-3	and	1e-4 respectively.
The above results demonstrate that the Gram regularization is more effective than $L_2$ regularization in the retrieval task.

\noindent \textbf{Influence of number of views.}
To investigate the effect of the number of views, 
we test our method on ModelNet40 with different numbers of rendered views.
The results are shown in Tab \ref{tab_view}.
As we can see, the performance first increases and then drops because of the addition of noise images.
The best performance is achieved with 8 views.
Therefore, we use 8 views for each shape in the testing.

\begin{table}
\begin{center}
\scalebox{1}[1]{
\begin{tabular}{|l|cc|}
\hline
\#View		& AUC 			& MAP\\
\hline
4			& 83.48\% 		& 82.20\% 		\\
6			& 83.42\% 		& 82.16\% 		\\	
8			& 85.02\% 		& 83.85\% 		\\
10			& 83.80\% 		& 82.57\% 		\\
16			& 83.15\% 		& 81.84\% 		\\
\hline
\end{tabular}
}
\end{center}
\caption{The retrieval results of different numbers of views on ModelNet40 dataset.}
\label{tab_view}
\end{table}

\noindent \textbf{Influence of learning rate.}
In this experiment, we explore how the learning rate $\epsilon$ affects the optimization process with Gram regularizer.
We vary $\epsilon$ from 3e-3 to 1e-4 in the training of our method and test the model by measuring the MAP on ModelNet40 dataset.
Other settings remain the same in this comparison experiment.
The comparison result is shown in Tab \ref{tab_lr}, the MAP first augments and then decreases that achieves peak with  $\epsilon$ 1e-3. 

\begin{table}
\begin{center}
\scalebox{1}[1]{
\begin{tabular}{|l|cccc|}
\hline
$\epsilon$		& 3e-3 			& 1e-3 	& 3e-4		& 1e-4\\
\hline
MAP			& 80.41\% 		& 83.85\% 	& 77.75\% & 72.02\% 	\\
\hline
\end{tabular}
}
\end{center}
\caption{The retrieval results of different learning rate on ModelNet40 dataset.}
\label{tab_lr}
\end{table}

\begin{figure}
\centering
\includegraphics[width=\linewidth]{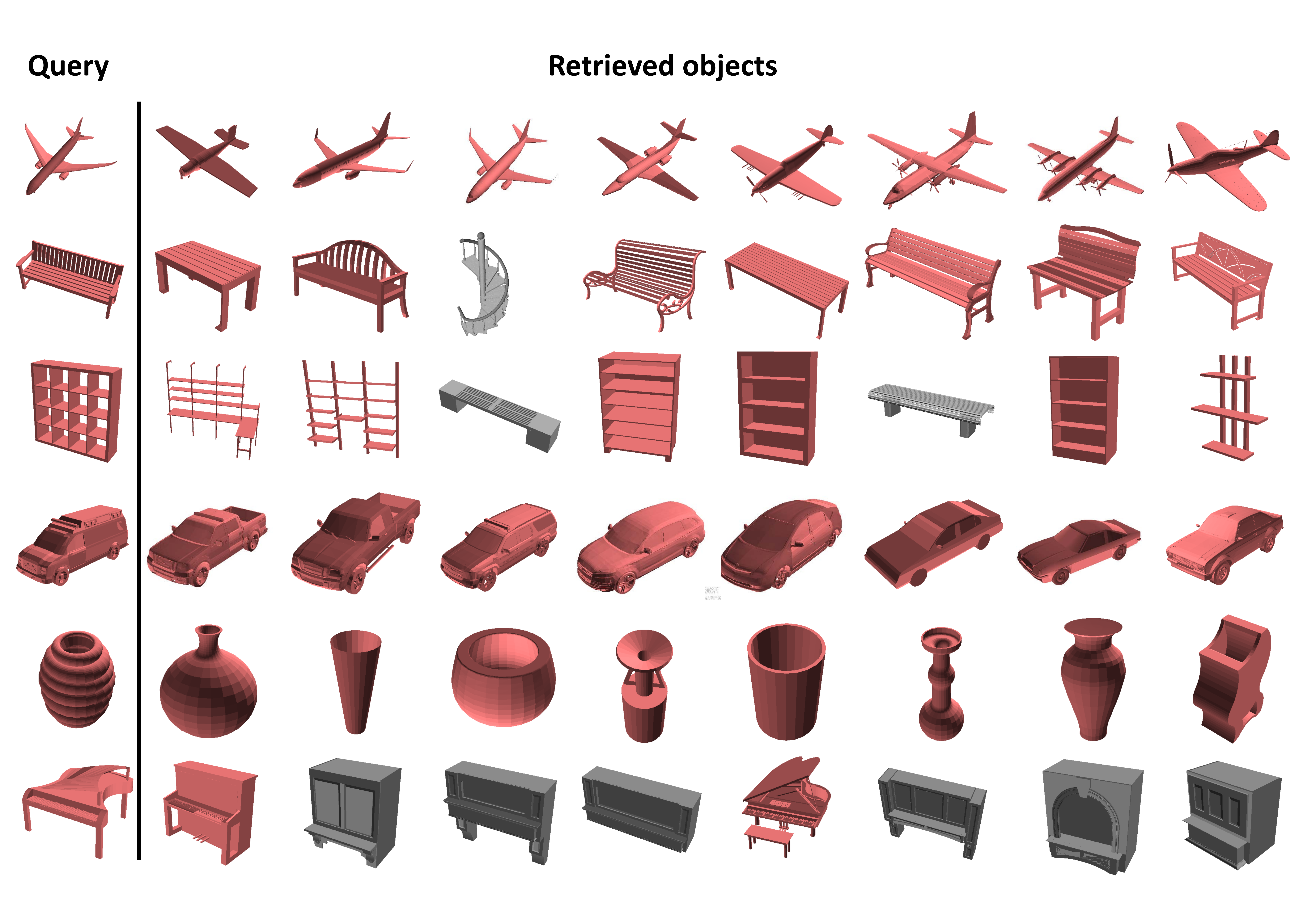}
\caption{Retrieval examples on ModelNet40 dataset. 
The querys are shown on the left column and each row places the first eight retrieved objects of corresponding query.
The objects in gray represent mistakes.}
\label{fig_vl}
\end{figure}

\noindent \textbf{Visualization of retrieval results.}
In Fig. \ref{fig_vl}, we show some examples of retrieved objects of our method. 
We can see that the retrieval performance is class-related. 
Some classes have high precision like Airplane while the last row is an example of difficult samples which are similar in appearance. 

%%%%%%%%%%%%%%%%%%%%%%%%%%%%%%%
\section{Conclusion}
In this paper, we propose a novel regularization method named Gram regularization to improve the generalization ability of neural network.
The proposed mothed encourage weight kernels to extract more information by decreasing their correlations.
The regularizer can be added into existing CNN architectures without bell and whistle and all the performances are reinforced. 
Abundant experimental results on two 3D shape datasets demonstrate the superiority of the learned 3D shape representations under Gram regularization. 
In the future, we would like to explore more on the correlation between weight and data for further research. 

\section*{Acknowledgement}
This work is supported by the Science,Technology and Innovation Commission of Shenzhen Municipality Foundation  (No.JCYJ20180307123632627), the Beijing Municipal Natural Science Foundation (No.L182014), and the National Natural Science Foundation of  China (No.61972014).

\section*{References}

\bibliography{GramReg}

\end{document}